%% file: root.tex
\documentclass[letterpaper, 10 pt, conference]{ieeeconf}  % Comment this line out if you need a4paper

\IEEEoverridecommandlockouts                              % This command is only needed if 
\usepackage{graphics} % for pdf, bitmapped graphics files
\usepackage{epsfig} % for postscript graphics files
\usepackage{amsmath} % assumes amsmath package installed
\usepackage{amssymb}  % assumes amsmath package installed
\usepackage{graphicx}
\usepackage{cite}
\usepackage{soul}
\usepackage{environ}
\usepackage{multirow}
\usepackage[table]{xcolor}
\usepackage{url}
\usepackage{color}

\usepackage{xcolor}
\usepackage{mathtools}
\usepackage{siunitx}
\usepackage{adjustbox}
\usepackage{float}
\usepackage[font=small,tableposition=top]{caption}
\usepackage{booktabs}

\usepackage{pgfplots}
\pgfplotsset{compat=newest} 
\usetikzlibrary{plotmarks}
\usepgfplotslibrary{patchplots}
\usepackage{grffile}
\usepackage{threeparttable}
\usepackage{gensymb}

\NewEnviron{scaletikzpicturetowidth}[1]{%
        \def\tikz@width{#1}%
        \begin{lrbox}{\measure@tikzpicture}%
        \BODY
        \end{lrbox}%
        \pgfmathparse{#1/\wd\measure@tikzpicture}% 
        \BODY
}
\makeatother

\graphicspath{{./}{./pics/}{../pics}}
\DeclareGraphicsExtensions{.jpg,.JPG,.png,.PNG,.pdf,.PDF,.eps,.EPS}

\usepackage[hidelinks]{hyperref}
\title{\LARGE \bf
iCaps: Iterative Category-level Object Pose and Shape Estimation}

\author{Xinke Deng$^{*1}$, %
Junyi Geng$^{*2}$, %
Timothy Bretl$^1$, %
Yu Xiang$^3$ %
and Dieter Fox$^{4,5}$
\thanks{*Xinke Deng and Junyi Geng contributed equally to this work, $^{1}$University of Illinois at Urbana-Champaign, $^{2}$Carnegie Mellon University, $^{3}$University of Texas at Dallas, $^{4}$University of Washington, $^{5}$NVIDIA}%
}

\begin{document}

\maketitle
\thispagestyle{empty}
\pagestyle{empty}

%%%%%%%%%%%%%%%%%%%%%%%%%%%%%%%%%%%%%%%%%%%%%%%%%%%%%%%%%%%%%%%%%%%%%%%%%%%%%%%%
\begin{abstract}
This paper proposes a category-level 6D object pose and shape estimation approach iCaps\footnote{The code is at \url{http://github.com/aerogjy/iCaps}. A video can be found at \url{https://youtu.be/3AMcM3uUaUw}.}, which allows tracking 6D poses of unseen objects in a category and estimating their 3D shapes. We develop a category-level auto-encoder network using depth images as input, where feature embeddings from the auto-encoder encode poses of objects in a category. The auto-encoder can be used in a particle filter framework to estimate and track 6D poses of objects in a category. By exploiting an implicit shape representation based on signed distance functions, we build a LatentNet to estimate a latent representation of the 3D shape given the estimated pose of an object. Then the estimated pose and shape can be used to update each other in an iterative way. Our category-level 6D object pose and shape estimation pipeline only requires 2D detection and segmentation for initialization. We evaluate our approach on a publicly available dataset and demonstrate its effectiveness. In particular, our method achieves comparably high accuracy on shape estimation.

% By exploiting the progress on implicit shape modeling, we combine the continuous signed distance function based shape representation with category-level pose estimation so that the pose and shape can be jointly estimated. In particular, we build a LatentNet to estimate the object shape and perform pose refinement based on the shape estimation. Our category-level 6D object pose and shape estimation pipeline only requires 2D detections for initialization. We compare our method with other category-level 6D object estimation methods on a publicly available dataset and demonstrate that our method is able to achieve state-of-the-art performance.
\end{abstract}

%%%%%%%%%%%%%%%%%%%%%%%%%%%%%%%%%%%%%%%%%%%%%%%%%%%%%%%%%%%%%%%%%%%%%%%%%%%%%%%%

\input{0_intro}

\input{1_related_work}

\input{2_methodology}

\input{3_experiments}

\section{Conclusion}

This paper proposes a category-level 6D object pose and shape estimation approach. We design an auto-encoder network for depth measurements so that the feature embeddings are independent of the object instances. We combine the continuous SDF based shape representation with the category-level pose estimation, where object pose and shape can be estimated and refined iteratively to improve each other. Our pose tracking only requires 2D detection for initialization. We evaluate our method on a publicly available dataset and demonstrate its effectiveness. In particular, our method achieves comparably high accuracy on shape estimation. Future work includes using the reconstructed shape to update the codebook and aggregating the point clouds from different view points to further improve the performance.

% \yu{Add 1 to 2 sentences for future work.}

% In particular, we build a LatentNet to estimate the object shape and perform the pose refinement based on the shape estimation.

%In this work, we propose a category-level 6D object pose and shape estimation framework based on Rao-Blackwellized particle filter with an implicit shape representation  (Fig. \ref{fig:category-title}). We develop a category-level auto-encoder network for depth measurements so that the feature embeddings are independent of the object instances. By exploiting the progress on implicit shape modeling, we combine the continuous SDF based shape representation with category-level pose estimation so that the pose and shape can be jointly estimated. In particular, we build a LatentNet to estimate the object shape and perform the pose refinement based on the shape estimation. Our category-level 6D object pose and shape estimation pipeline only requires 2D detections for initialization, which can be provided by any detectors such as \cite{girshick2015fast, redmon2016you}. 

%\section*{ACKNOWLEDGMENT}

\bibliographystyle{IEEEtran}
% \bibliography{IEEEabrv, } 

\bibliography{references}

\end{document}

%% file: 0_intro.tex
\section{Introduction}
Estimating 6D object poses, i.e., 3D translations and 3D orientations of objects with respect to cameras, is crucial for a variety of real-world applications ranging from robotic navigation and manipulation to augmented reality and virtual reality. The majority of existing works have so far mainly dealt with the \textit{instance-level} 6D pose estimation \cite{xiang2017posecnn, li2017deepim, sundermeyer2018implicit, wang2019densefusion, peng2019pvnet, labbe2020, deng2021pose}, where a set of 3D CAD models of known instances are given as priors. The problem is thereby reduced to finding the sparse or dense correspondence between a target object and a prior 3D model. Although 3D CAD models are available in some industrial applications such as assembling different parts, the requirement still significantly limits many practical robotic applications since it can be expensive or even impossible to acquire 3D CAD models of all the objects in an environment.

% This limitation motivate leads to the generalization from instances to categories.

In \textit{category-level} 6D object pose estimation, a target object may be unseen during training and its 3D CAD model is not available. While even in the same category, objects can exhibit significant differences in color, texture, shape and size. Without 3D CAD models, the correspondence matching approaches would run into a significant challenge under the considerable shape variations among objects. Therefore, the major challenge of category-level pose estimation is how to handle the intra-class variability~\cite{sahin2019instance}. Some recent progress has been made to address this challenge\cite{chen2020category,yen2020inerf}. Wang \textit{et al.}~\cite{wang2019normalized} propose to transform every object pixel to a canonical 3D space as keypoints for pose estimation. Although showing significant progress, this method still requires handling symmetric objects separately and is susceptible to noises from clutter and occlusion. Due to the difficulty in learning dense correspondences between pixels and 3D coordinates, in \cite{chen2020learning}, the authors propose to learn a canonical shape shape with RGB-D fusion features. More recently, FS-net~\cite{chen2021fs} was proposed  to directly predict object poses and sizes where an orientation-aware autoencoder with 3D graph convolution is used for latent feature extraction. However, these methods focus on providing pose estimation for a single image and cannot take advantage of the temporal consistency among video frames, which is commonly observed in robotic tasks. Very recently, significantly process has been made in category-level pose tracking~\cite{wang20196, weng2021captra, wen2021bundletrack}. While these methods successfully exploit the temporal consistency and improve pose accuracy compared to single-frame methods, they usually require ground truth 6D object pose for initialization. %\yu{Can we say they use ground truth 6D pose for initialization?} %, \yu{project images to 3D? Should it be pixels?}
 %to use learned neural networks to compute a category-level and view-factorized RGB-D embedding, where 6D pose is regressed through a separate deep neural network by implicitly contrasting embeddings with view-dependent features

\begin{figure}
    \centering
    \includegraphics[width=0.5\textwidth]{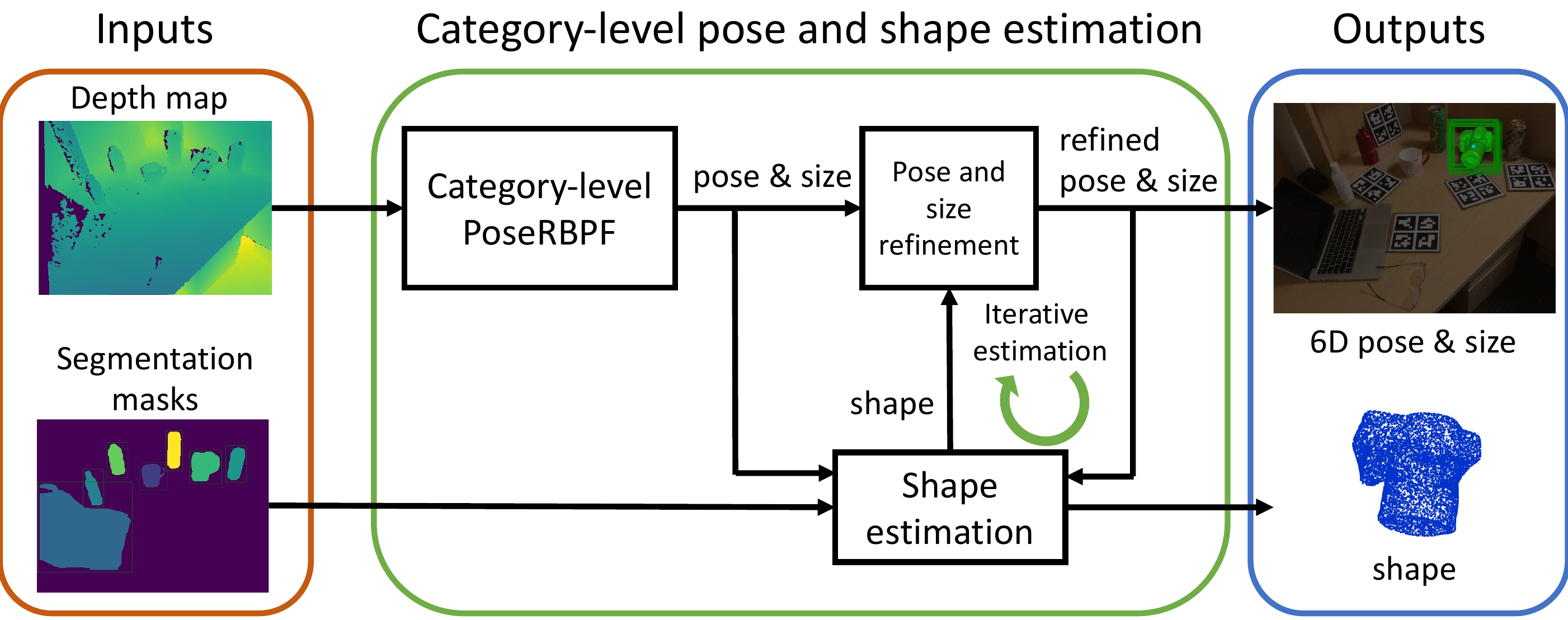}
    \caption{The proposed system takes in depth maps and segmentation masks as inputs, and jointly estimates the 6D poses, size, and shape of the unseen objects in an iterative way so that pose and shape can improve each other.}% \yu{Let's highlight the iterative refinement of pose and shape in this figure. The main point for intro is not only we can estimate shape and pose jointly, but pose and shape can improve each other iteratively. We need to highlight this.}
    \label{fig:category-title}
    \vspace{-6mm}
\end{figure}

While more attention has been made in improving 6D pose estimation, object shape estimation, however, has been less explored. Good shape estimation can intuitively assist object pose estimation especially for category-level pose estimation where the intra-class variability is the major challenge~\cite{shi2021optimal}. Existing works for shape estimation either represent object shape with a canonical shape and size~\cite{chen2020learning} or use reconstructed point clouds from an auto-encoder network~\cite{tian2020shape, chen2021fs}. However, in these works, only the decoder is retained for inference which makes the framework less efficient since the trained encoder is unused at test time.  In such situations, learning implicit representations of object shapes opens another possibility for shape estimation. In \cite{park2019deepsdf}, the authors propose continuous Signed Distance Function (SDF) for shape representation. The decision boundary of the learned shape-conditioned classifier is the surface of the shape itself. The developed probabilistic auto-decoder only model can produce high quality continuous surfaces with complex topology for canonical input point clouds.
%\yu{I don't get this. Why not using the encoder is a bad thing?}
%\yu{Is there a classifier in DeepSDF?}

In this work, we propose a category-level 6D object pose and shape estimation framework based on a Rao-Blackwellized particle filter with an implicit shape representation  (Fig.~\ref{fig:category-title}). Built upon our previous instance-level pose estimation work PoseRBPF~\cite{deng2021pose}, we develop a category-level auto-encoder network using depth measurements as input. Feature embeddings from the auto-encoder encode object poses while being independent of specific object instances. In addition, we combine the continuous SDF based shape representation with our category-level pose estimation so that object pose and shape can be jointly estimated. In particular, we build a LatentNet to estimate the object shape based on the current pose estimation. Then we can refine the pose using the estimated shape. In this way, both the pose and the shape of an object can improved in an iterative refinement fashion. Different from previous methods~\cite{wang20196, weng2021captra, wen2021bundletrack}, our category-level 6D object pose and shape estimation pipeline only requires 2D detection and segmentation for initialization, which can be provided by any object detector such as \cite{girshick2015fast, he2017mask}. 

Through evaluating our framework on a category-level pose estimation benchmark~\cite{wang2019normalized}, we show the effectiveness of our approach. In particular, our method achieves comparably high accuracy on shape estimation. To summarize, the main contributions of this paper are as follows:
\begin{itemize}
    \item We propose a category-level 6D object pose estimation approach based on a Rao-Blackwellized particle filter. The devised category-level auto-encoder network uses depth measurements as input so that it can better handle the intra-class variability.
    \item We propose LatentNet, a deep neural network which directly predicts the implicit SDF-based shape representation of the object.
    %a category-level 3D object shape estimation method with an implicit SDF-based shape representation. The proposed LatentNet in our method makes shape inference more efficient.
    \item We combine the SDF-based shape representation with the category-level pose estimation so that pose estimation and shape estimation can benefit and improve each other in an iterative way. 
\end{itemize}

%% file: 1_related_work.tex
\section{Related Work}
\textbf{Instance-level pose estimation}.
The majority of object pose estimation work mainly deal with instance-level pose estimation \cite{xiang2017posecnn, li2017deepim, sundermeyer2018implicit, wang2019densefusion, peng2019pvnet, labbe2020, deng2021pose}. Traditionally, template matching \cite{collet2011moped} and local feature matching \cite{hinterstoisser2012model,cao2016real} are used to tackle the problem. In recent years, learning-based methods have received more attention due to their ability to generalize to different backgrounds, lighting, and occlusions \cite{brachmann2014learning,krull2015learning}. For instance, Kehl et al. \cite{kehl2017ssd} extend the SSD detection network \cite{liu2016ssd} to 6D pose estimation by adding viewpoint classification to the network. Tekin et al. \cite{tekin2018real} utilize the YOLO architecture \cite{redmon2016you} to detect 3D bounding box corners of objects in images, and then the 6D pose is recovered by solving the PnP problem. PoseCNN \cite{xiang2017posecnn} designs an end-to-end network for 6D object pose estimation based on the VGG architecture \cite{simonyan2014very}. Sundermeyer \textit{et al.}~\cite{sundermeyer2018implicit} introduce an implicit way of representing 3D rotations by training an auto-encoder for image reconstruction, which does not need to pre-define the symmetry axes for symmetric objects. Significant progress has also been make to track the 6D pose of known objects. In \cite{deng2021pose}, a Rao-Blackwellized particle filter is combined with a learned auto-encoder network for tracking the probability distribution of 6D poses. In \cite{wense3tracknet}, the authors propose to represent the incremental motion with Lie Algebra and predict it by comparing the rendered object and the RGB-D image in the feature space. The main limitation of instance-level pose estimation is that 3D models of exactly the same object instances are required. In many applications, it is difficult to obtain these 3D models.

%The goal is to estimate the 3D translation and 3D orientation for objects with a set of known instances as prior.
%Detecting 3D bounding box corners or object key points for 6D object pose estimation is also explored in \cite{tremblay2018corl:dope,marion2018label,peng2019pvnet,song2020hybridpose}.
% Known 3D object models are usually available for training and testing. There are in general three types of instance-level pose estimation methods based on the 3D model: template matching based methods; correspondence matching based methods and voting based methods. Many works on instance-level object pose estimation pursue template matching based methods. These methods try to align the template generated from the object CAD model in various poses to the observed input images or depth map through hand-designed or learned feature descriptors. The object pose is retrieved from the template with the best match or obtained from the 3D model registration. Correspondences matching based methods aims to match the target object to the corresponding 3D model, where 2D-3D correspondences, or 3D-3D correspondences are established. Then, the pnp and SVD problem with 2D-3D correspondences, or 3D-3D correspondences are solved respectively. Several other works also generate voting candidates based on these established correspondences. Best candidate is then selected based using the RANSAC algorithm. However, all those methods are based on the known 3D object models that are not available for the category-level object pose estimation. 

\textbf{Category-level pose estimation.}
The limitation on the CAD model requirement motivates development of \textit{category-level} pose estimation methods. There are efforts on category-level 3D detection~\cite{qi2018frustum, xiang2014monocular, MousavianCVPR17}, where the task is to predict 3D bounding boxes of objects. However, 3D bounding boxes are not sufficient for applications such as robotic manipulation. Some recent progress has been made to address the problem of category-level 6D pose estimation~\cite{wang2019normalized, wang20196, chen2020learning, chen2021fs, weng2021captra, wen2021bundletrack}. Besides the mentioned work on category-level pose estimation for single frames such as NOCS~\cite{wang2019normalized}, CASS~\cite{chen2020learning}, FS-Net~\cite{chen2021fs} and the category-level tracking work 6-PACK~\cite{wang20196}, recently, Weng~\textit{et al.} \cite{weng2021captra} propose a unified framework that can handle 9DoF category-level pose tracking for rigid object instances. More recently, Wen \textit{et al.} propose BundleTrack~\cite{wen2021bundletrack} which uses deep neural networks to extract and match keypoints, then pose graph optimization is used for pose tracking. In contrast to the mentioned tracking systems which require ground truth 6D pose of objects to initialize the tracking process, our system is able to initialize pose tracking with only 2D detection and segmentation of objects which are easier to obtain using state-of-the-art object detection and segmentation methods such as \cite{he2017mask}.

\begin{figure*}
    \centering
    \includegraphics[width=0.72\textwidth]{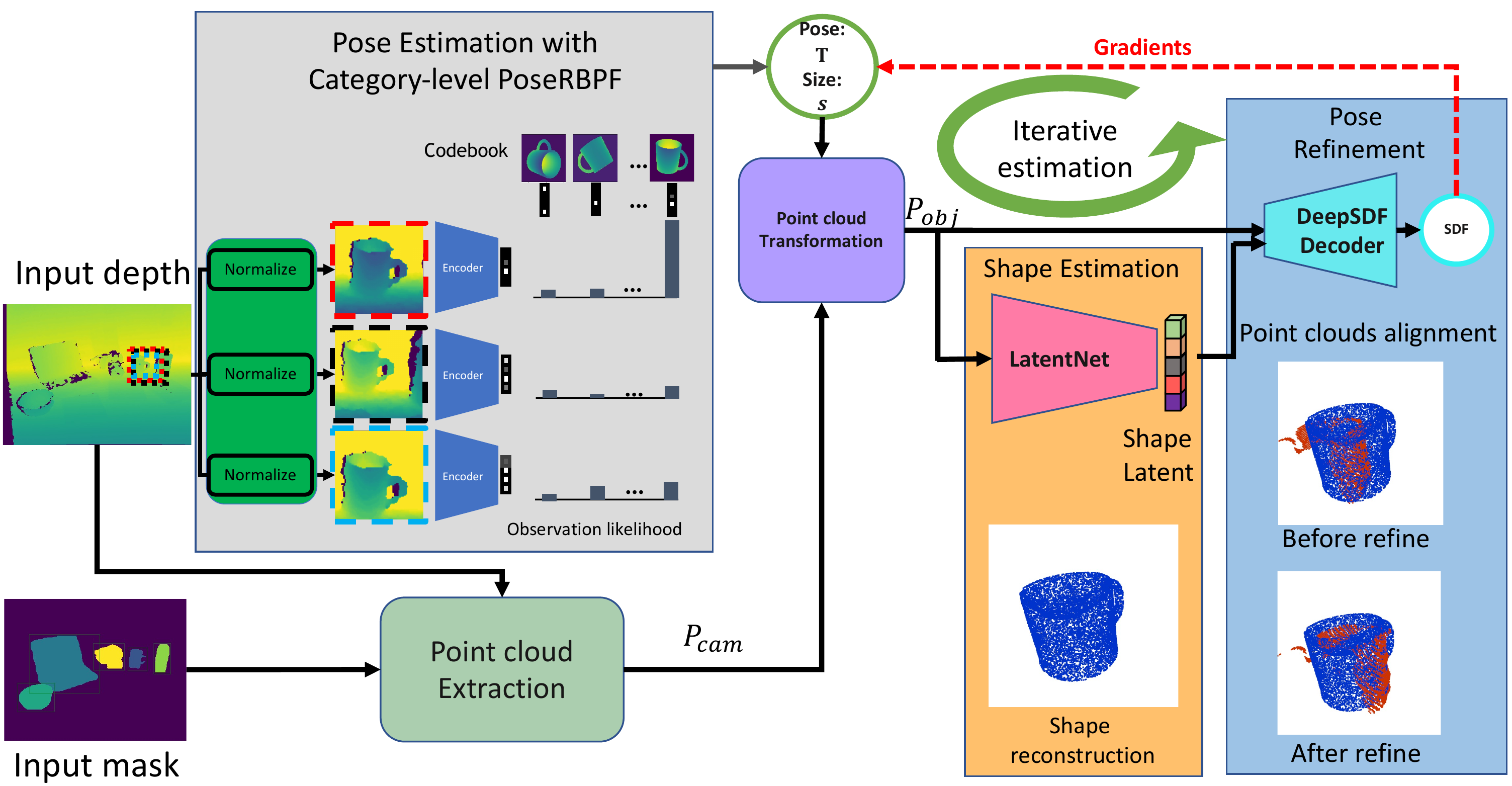}
    \caption{
    Architecture the proposed category-level pose and shape estimation framework. We first use a category-level generalization of the PoseRBPF system to estimate the 6D pose and the size of the object. An implicit shape representation (shape latent) is predicted with a deep neural network (LatentNet). The 6D pose and size of the object can be further refined by minimizing the SDF computed according to the predicted shape latent and a deep auto-decoder network.}
    \label{fig:system-diagram}
    \vspace{-6mm}
\end{figure*}

\textbf{Object shape estimation.}
% shape representation. implicit representation (deepsdf etc). how to estimate the shape? what's the role of pose
Object shape estimation aims to infer 3D shape of objects given partial or sparse observations.
Traditionally, surface reconstruction methods are usually adopted such as completing point clouds into dense surfaces \cite{carr2001reconstruction, kazhdan2013screened}. However, only single object shape can be modeled at one time by these methods rather than a dataset as a batch. %\yu{what does only single shape mean?} 
Various recent works use data-driven approaches for the 3D shape estimation task. Most of these works pursue the auto-encoder architectures to encoder the inputs of RGB images \cite{stutz2018learning}, depth maps \cite{rock2015completing}, point clouds \cite{tian2020shape}, occupancy voxels \cite{wu20153d} or discrete SDF voxels \cite{dai2017shape} into a latent vector and subsequently reconstruct the full shape prediction. In these works, however, only the decoder is retained for inference which makes the framework less efficient since the trained encoder is unused at test time. 
Recently, learning implicit representations of object shapes attracts more attention \cite{park2019deepsdf, Mescheder2019CVPR}. For instance, Park \textit{et al.} \cite{park2019deepsdf} propose continuous SDF for shape representation. The decision boundary of the learned shape-conditioned classifier is the surface of the shape itself. Although the methods can produce high quality continuous surfaces with complex topology, they are usually view dependent and assume ground truth 6D pose to convert the input point cloud to a canonical view before performing shape estimation. In this paper, we estimate the object 6D pose and the shape jointly because they are tightly correlated: good shape estimation improves pose estimation while accurate pose simplifies the shape estimation.

% the shape estimation task usually assumes the input point cloud is already in a canonical frame, i.e., where the pose of the object is accurate. In this paper, we exploit this progress on implicit shape modeling, and combine it with category-level pose estimation so that the pose and shape can be jointly estimated and improve each other.

% Several works spent effort on object shape estimation while estimating the category-level pose to handle the \textit{intra-class variation}. CASS \cite{chen2020learning} learns a canonical shape space as the unified representation. A variational auto-encoder (VAE) is trained in a cross-category fashion for generating 3D point clouds in the canonical space from an RGBD image. The VAE model is able to reconstruct the 3D point cloud of the target object. Tian \textit{et al.} \cite{tian2020shape} design an autoencoder that trains on a collection of object models and compute the mean latent embedding for each category to learn the categorical shape priors. A deep network is then trained to reconstruct the 3D object model by explicitly modeling the deformation from the pre-learned categorical shape prior. FS-Net \cite{chen2021fs} proposes a 3D graph convolution (3DGC) autoencoder to extract the point cloud shape feature. Observed point cloud reconstruction can be obtained. 

%% file: 2_methodology.tex
\section{Category-level Pose and Shape Estimation}
The architecture of our category-level pose and shape estimation framework iCaps is shown in Fig. \ref{fig:system-diagram}. In this section, we first formulate the problem. Then, we describe in details how to generalize PoseRBPF \cite{deng2021pose} from instance-level to category-level for 6D pose and size estimation. Subsequently, we illustrate how to use a deep neural network (LatentNet) to predict an implicit representation of the object shape (shape latent). Finally, we show that the predicted shape latent can be used together with an auto-decoder network proposed in \cite{park2019deepsdf} to further refine the 6D pose.

\subsection{Problem Formulation}
% \begin{itemize}
%     \item Input: depth, masks
%     \item define estimates: 6DoF pose, size, and shape.
% \end{itemize}
The goal of category-level 6D object pose and shape estimation is to estimate the 3D rotation $\mathbf{R}$, 3D translation $\mathbf{T}$, together with the object size $s$ and shape $\mathbb{S}$ of an object given depth measurements $\mathbf{Z}$ and instance segmentation mask $\mathbf{\hat{\Omega}}$. The size $s$ represents the metric length of the longest distance of the object 3D bounding box. The segmentation mask $\mathbf{\hat{\Omega}}$ can be estimated with methods such as Mask R-CNN \cite{he2017mask}.

\subsection{A Rao-Blackwellized Particle Filter for Category-level 6D Pose Estimation}
\subsubsection{Particle Filtering Formulation}
At time step $k$, our first task is to estimate the posterior distribution $P(\mathbf{R}_k, \mathbf{T}_k, s_k|\mathbf{Z}_{1:k})$ of 3D rotation $\mathbf{R}_k$, 3D translation $\mathbf{T}_k$ and object size $s_k$ given observations $\mathbf{Z}_{1:k}$. Similar to \cite{deng2021pose}, we factorize the 6D pose and size estimation problem into 3D rotation estimation and 3D translation and size estimation since the translation and object size determine the center and scale of the object in the image, and the 3D rotation can be estimated based on the measurements in the Region of Interest (RoI) of the object. Therefore, the posterior is decomposed as:
\begin{equation}
    P(\mathbf{R}_k, \mathbf{T}_k, s_k|\mathbf{Z}_{1:k}) = P(\mathbf{T}_k, s_k|\mathbf{Z}_{1:k})P(\mathbf{R}_k|\mathbf{T}_k, s_k, \mathbf{Z}_{1:k})
\end{equation}
where $P(\mathbf{T}_k, s_k|\mathbf{Z}_{1:k})$ encodes the RoI of the object, and $P(\mathbf{R}_k|\mathbf{T}_k, s_k, \mathbf{Z}_{1:k})$ models the rotation distribution conditioned on the RoI and the observations.

According to the above factorization, the particles in the Rao-Blackwellized particle filter can be represented as $\mathcal{X}_{k} = \{ \mathbf{T}_{k}^i, s_{k}^i, P( \mathbf{R}_{k}|{\mathbf{T}_{k}^i, \mathbf{Z}_{1:k}}), w_{k}^i \}_{i=1}^N$ with $N$ the number of the particles. Here, $\mathbf{T}_{k}^i$ and $s_k^i$ denote the translation and size of the $i$th particle. $P( \mathbf{R}_{k}|{\mathbf{T}_{k}^i, s_k^i, \mathbf{Z}_{1:k}})$ denotes the discrete \emph{distribution} of the particle over the object rotation conditioned on the translation, the size and the images, and $w_{k}^i$ is the importance weight of the particle. The 3D rotation is discretized on azimuth, elevation and in-plane rotation with bins of size 5 degrees, resulting in 191,808 bins for each particle. According to the particle filtering formulation, the importance weight $w_k^i$ for sampling translation and size can be computed as:
\begin{equation} \label{eq:weight_c}
    w_k^i   \propto P(\mathbf{Z}_k|\mathbf{T}_k^{i}, s_k^i) 
            \approx \sum_j  P(\mathbf{Z}_k|\mathbf{T}_k^i, s_k^i, \mathbf{R}_k^j)P(\mathbf{R}_k^j), 
\end{equation}
where $\mathbf{R}_k^j$ represents the $j$th discretized rotation.

\subsubsection{Category-level Auto-encoder}
In \cite{deng2021pose}, a learned auto-encoder network is used to map the input image inside the RoI to a synthetic image of the object with the same pose. Since the lighting is constant and there is no background and occlusion in the synthetic image, the embedding from the encoder only depends on the object's appearance in different orientations. The rotation distribution can thus be inferred by comparing the embedding with a pre-computed codebook.

\begin{figure}
    \centering
    \includegraphics[width=0.3\textwidth]{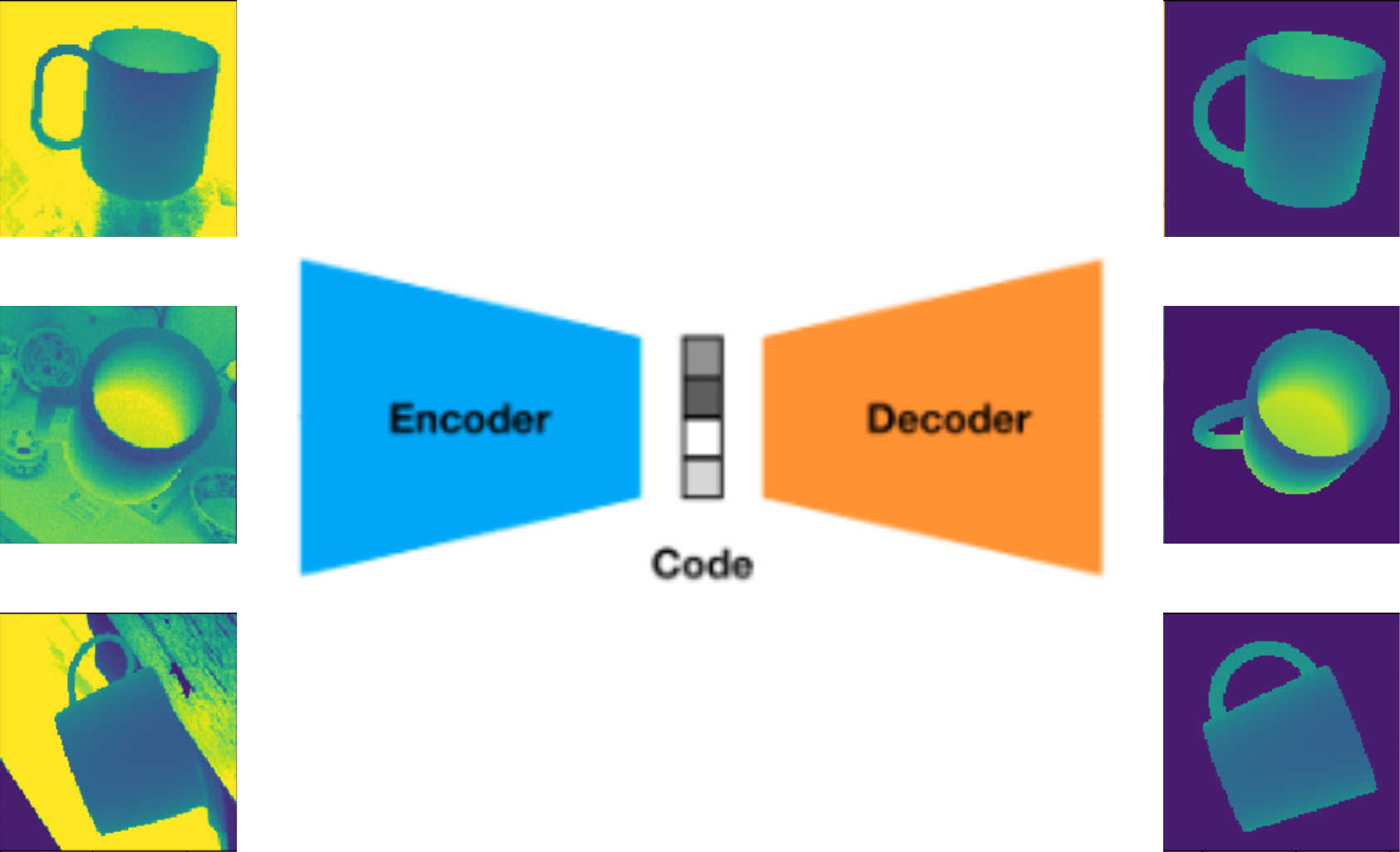}
    \caption{The category-level auto-encoder for normalized depth. The auto-encoder takes in the normalized depth of an arbitrary object in the category and tries to reconstruct the normalized depth of a canonical object so that the embedding is robust to the variance between object instances.}
    \label{fig:category-aae}
    \vspace{-6mm}
\end{figure}

To deal with the challenge of ``intra-class variation''~\cite{sahin2019instance} , we propose to train the auto-encoder to take in the rendered depth of an arbitrary object in the category and reconstruct the rendered depth of a canonical object so that the embedding is invariant to differences between object instances in the category (Fig. \ref{fig:category-aae}). 

We take advantage of ShapeNet \cite{shapenet2015}, which is a large scale database of 3D CAD models. For each category, we manually select a representative object among the models as the canonical object. The dimensions of ShapeNet models are normalized with the object size, which is defined as the diagonal length of the 3D bounding box. We render an arbitrary object in the category with a canonical translation $\mathbf{T}_0=(0, 0, z_0)^T$ and rotation $\mathbf{R}$ which is uniformly sampled in the rotation space. Denoting the rendered depth map as $\mathbf{D}_i$, the normalized depth map $\mathbf{\bar{D}}_i$ is computed as $\mathbf{\bar{D}}_i = f_c((\mathbf{D}_i - z_0)/s + 0.5)$,
% \begin{equation}
%     \mathbf{\bar{D}}_i = f_c((\mathbf{D}_i - z_0)/s + 0.5), \label{eq:train_norm}
% \end{equation}
where $f_c(x)$ is a clamping function defined as $f_c=\max(0, \min(1, x))$ so that the values are between 0 and 1, and $s=1$. We also render the canonical object with the same pose and result in normalized depth map $\mathbf{\bar{D}}_o$. The auto-encoder is trained to take in $\mathbf{\bar{D}}_i$ and reconstruct $\mathbf{\bar{D}}_o$ so that the embeddings are instance-agnostic, i.e., every instance is mapped to the canonical object. After training the auto-encoder, we can compute the codebook by encoding the normalized depth map of the canonical object with 6D pose $(\mathbf{T}_0, \mathbf{R}^j)$, where $\mathbf{R}^j$ is a discrete orientation in $\mathbf{SO}(3)$.

\subsubsection{Observation Likelihoods Computation}
\begin{figure}
\centering
    \includegraphics[width=0.35\textwidth]{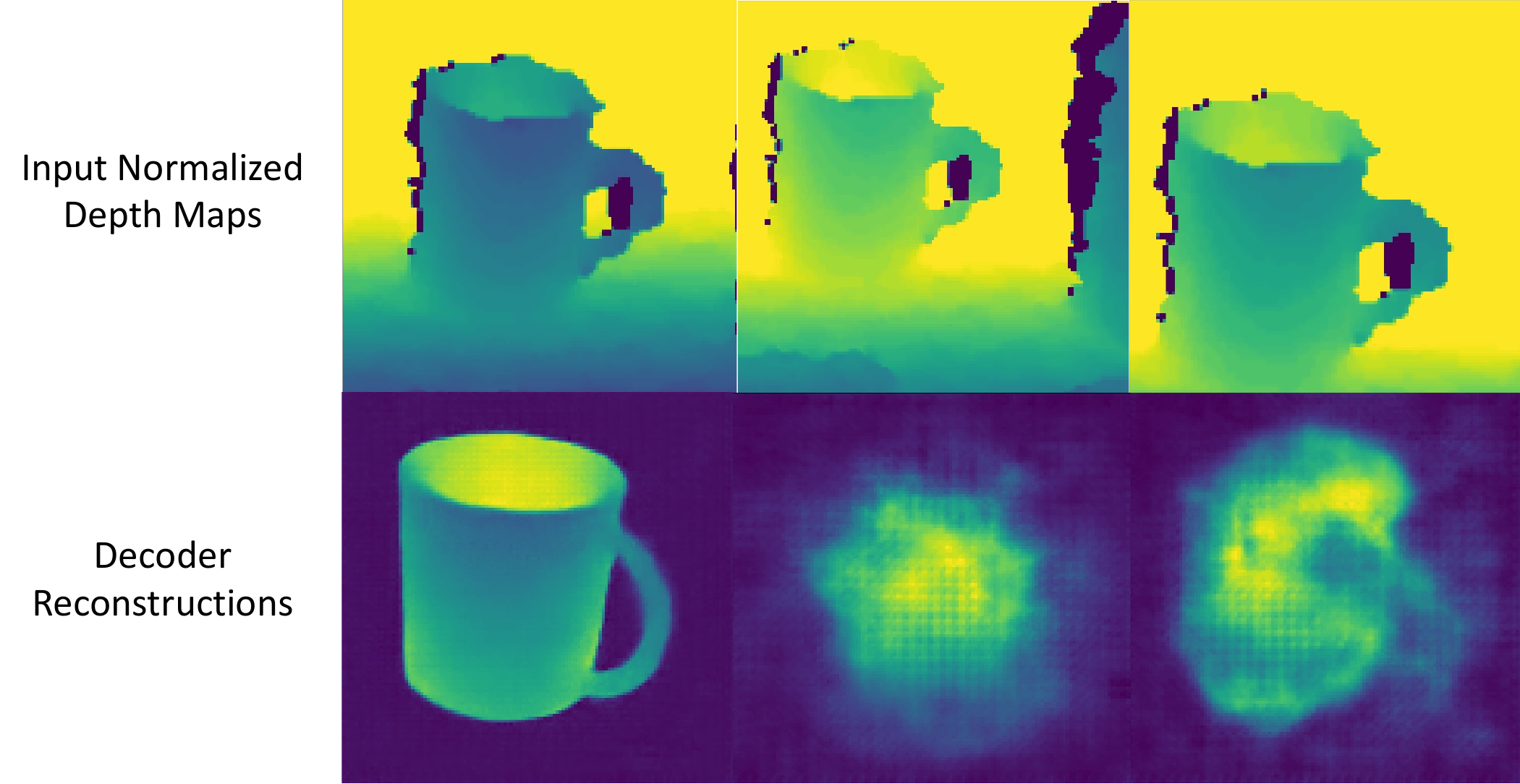}
    \caption{Reconstruction of the normalized depth maps with the auto-encoder. Left column shows the reconstruction from ground truth translation and size. The rest two columns are with incorrect translation and size. As it is shown, incorrect translation and size estimation will cause the bad reconstruction and therefore low similarity to the codebook. This property makes the auto-encoder a suitable choice for computing the observation likelihood.}
    \label{fig:reconstruction}
    \vspace{-6mm}
\end{figure}

As shown in Fig. \ref{fig:reconstruction}, when the translation and the size are incorrectly sampled, the auto-encoder cannot reconstruct meaningful normalized depth maps from the inputs. Therefore the similarities between the embedding and codes in the codebook will be low. We exploit this property to compute observation likelihoods. The process of computing observation likelihoods is shown in Fig. \ref{fig:system-diagram}. We first compute the RoI based on the 3D translation and size hypothesis $(\mathbf{T}_k, s_k)$. The center of the RoI $(u_k, v_k)$ is computed by projecting the 3D translation $\mathbf{T}_k = (x_k, y_k, z_k)^T$ according to the camera intrinsic parameters.
% \begin{equation} \label{eq:projection}
%     \begin{bmatrix}
%     u_k \\[0.5em] v_k
%     \end{bmatrix} = \begin{bmatrix}
%     f_x x_k/z_k + p_x \\[0.5em]
%     f_y y_k/z_k + p_y
%     \end{bmatrix},
% \end{equation}
% where $f_x$ and $f_y$ indicate the focal lengths of the camera, and $(p_x, p_y)^T$ is the principal point. 
The size of the RoI is determined by $\frac{z_k}{z_0}s_k$, where $z_0$ is the z coordinate of the canonical translation used in training the auto-encoder. Denoting the input depth image as $\mathbf{Z}_k^D$ and the corresponding part inside the RoI as $\mathbf{D}_k$, the normalized depth map $\mathbf{\bar{D}}_k$ can be computed as $\mathbf{\bar{D}}_k = f_c((\mathbf{D}_k - z_k)/s_k + 0.5)$
% \begin{equation}
%     \mathbf{\bar{D}}_k = f_c((\mathbf{D}_k - z_k)/s_k + 0.5),
% \end{equation}
to regularize the depth value between 0 and 1. The normalized depth map $\mathbf{\bar{D}}_k$ is fed into the auto-encoder to compute the feature embedding $\mathbf{c} = f(\mathbf{\bar{D}}_k)$. The observation likelihood $P(\mathbf{Z}_k|\mathbf{T}_k, s_k, \mathbf{R}_c^j)$ can be computed 
\begin{equation}
    P(\mathbf{Z}_k|\mathbf{T}_k, s_k, \mathbf{R}_c^j) = \phi\Big(\frac{\mathbf{c} \cdot f(\mathbf{\bar{D}}(\mathbf{R}_c^j, \mathbf{T}_0))}{\|\mathbf{c}\| \cdot \|f(\mathbf{\bar{D}}(\mathbf{R}_c^j, \mathbf{T}_0))\|}\Big), \label{eq:category_likelihood}
\end{equation}
where $\mathbf{R}_c^j$ is one of the discretized rotations in the codebook, $ f(\mathbf{\bar{D}}(\mathbf{R}_c^j, \mathbf{T}_0))$ is the code of the rotation in the codebook and $\phi(\cdot)$ is a Gaussian probability density function centered at the maximum cosine distance among all the codes in the codebook for all the particles. The probabilistic distribution of all the rotations in the codebook given a translation and a size can be computed according to Bayes' rule:
\begin{equation} 
    P(\mathbf{R}_c^j|\mathbf{T}_k, \mathbf{Z}_k, s_k) \propto P(\mathbf{Z}_k|\mathbf{T}_k, s_k, \mathbf{R}_c^j) \label{eq:rot_likelihood_cat}.
\end{equation} 

\subsubsection{Category-level 6D Object Pose Tracking}
% \begin{itemize}
%     \item initialization
%     \item tracking with 2D detection / processing model
%     \item algorithm
% \end{itemize}
Our category-level 6D object pose tracking pipeline can be initialized with any 2D object detector such as \cite{girshick2015fast, redmon2016you}. The 2D center of an object $(u_k, v_k)$ and the depth of a particle $z_k$ are sampled in the same way as~\cite{deng2021pose}. The object size $s_k$ is sampled as $s_k \sim \mathcal{U}\big(s_0-\Delta s/2, s_0+\Delta s/2)$,
%Denoting the center of the detected bounding box as $(\hat{u}, \hat{v})$, the object's center $(u_i, v_i)$ , depth $z_i$ and size $s_i$ are sampled as:
% \begin{equation}
%     %(u_i, v_i) &\sim \mathcal{N}\big((\hat{u}, \hat{v}), \sigma_{u}, \sigma_{v}\big) \nonumber\\ 
%     %z_i &\sim \mathcal{U}\big(\mathbf{Z}^D(u_i, v_i)-\frac{d}{2}, \mathbf{Z}^D(u_i v_i)+\frac{d}{2}) \nonumber\\
%     s_k \sim \mathcal{U}\big(s_0-\Delta s/2, s_0+\Delta s/2),
% \end{equation}
where $s_0$ is a constant denoting a prior of the object size for objects in the category, and $\Delta s$ denotes the range of the uniform distribution $\mathcal{U}$. 
%where $\sigma_{u}, \sigma_{v}$ denote the standard deviations of the Gaussian distributions for object center; $d$ and $\Delta s$ denote the range of a uniform distribution where $z_i$ and $s_i$ are sampled from respectively; and $s_0$ denotes the prior of the object's size. $\mathbf{Z}^D(u_i, v_i)$ is the depth value at pixel location $(u_i, v_i)$. 
%The 3D translation $\mathbf{T}^i$ can be computed by backprojecting the object center. %We then evaluate the likelihood of each hypothesis according to Eq.~\eqref{eq:weight_c}. The translation with the highest likelihood and the corresponding rotation distribution are used as the initial value of the particles.

In the following frames, the translation $\mathbf{T}_k$, size $s_k$, and rotation $\mathbf{R}_k$ are propagated based on a motion prior:
\begin{align}
    P(\mathbf{T_k}|\mathbf{T}_{k-1}, \mathbf{T}_{k-2}) &= \mathcal{N}(\mathbf{T}_{k-1} + \alpha(\mathbf{T}_{k-1} - \mathbf{T}_{k-2}), \mathbf{\Sigma}_{\mathbf{T}}), \nonumber\\
    P(s_k|s_{k-1}) & = \mathcal{N}(s_{k-1}, \sigma_s), \nonumber \\
    P(\mathbf{R}_k|\mathbf{R}_{k-1}) &= \mathcal{N}(\mathbf{R}_{k-1}, \mathbf{\Sigma}_{\mathbf{R}}),
\end{align}
where $\mathcal{N}(\cdot)$ represents a Gaussian distribution, $\Sigma_{\mathbf{T}}$, $\sigma_s$ and $\Sigma_\mathbf{R}$ represents the covariances of the corresponding Gaussian distributions, and $\alpha$ is a constant for the linear velocity model. Then the rotation is updated according to Eq.~\eqref{eq:rot_likelihood_cat}. The weight of each particle is computed according to Eq.~\eqref{eq:weight_c} and used to sample particles. The pose and size estimation can be obtained for every frame by computing the mean of the sampled particles.
%where $\mathcal{N}(\cdot)$ represents Gaussian distribution, and $\mathbf{\Sigma}_{\mathbf{T}}, \sigma_s, \mathbf{\Sigma}_{\mathbf{R}}$ represent the covariance for translation, size and rotation respectively. Then the rotation is updated according to Eq.~\eqref{eq:rot_likelihood_cat}. The weight of each particle is computed according to Eq.~\eqref{eq:weight_c} and then used to sample particles.

%The rotation distribution is updated with:
% \begin{equation}
%     P(\mathbf{R}_k|\mathbf{T}_k^i, s_k^i, \mathbf{Z}_{1:k}^i) = \eta P(\mathbf{R}_k|\mathbf{T}_k^i, s_k^i, \mathbf{Z}_k)P(\mathbf{R}_k|\mathbf{R}_{k-1})P(\mathbf{R}_{k-1})
% \end{equation}
% where $P(\mathbf{R}_k|\mathbf{T}_k^i, s_k^i, \mathbf{Z}_k)$ is computed according to Eq.~\eqref{eq:rot_likelihood_cat}. The weight of each particle is computed according to Eq.~\eqref{eq:weight_c} and then used to sample particles.
%To handle the abrupt camera motion, the translation of half of the particles is sampled according to the detection center.
\subsection{Category-level Object Shape Estimation}
%\xinke{maybe start with how is the shape represented? then explain why normalized point cloud is used.}

We design a point cloud based deep neural network named LatentNet to predict the objects' 3D shape (see Fig. \ref{fig:system-diagram}). The network takes as input the normalized 3D points from the depth image with respect to the object. In this way, we only consider the object shape information in LatentNet by transforming the point cloud using the estimated object pose. 

%\begin{definition} (Normalized point cloud)
    Normalized point cloud with respect to the object is defined as the product of the inverse transformation of the point cloud in camera frame $\mathbf{P}_{obj}$ with pose ($\mathbf{T}_k$, $\mathbf{R}_k$) and size $s_k$, namely,
    \begin{align}
        \bar{\mathbf{P}}_{\mathrm{obj}} = g(\mathbf{P}_{obj}, \mathbf{T}_k, \mathbf{R}_k, s_k) =\mathbf{R}_k^{-1} (\mathbf{P}_{\mathrm{obj}} - \mathbf{T}_k)/s_k. \label{eq:point_normalize}
    \end{align}
%\end{definition}
%&= g(\mathbf{P}_{obj}, \mathbf{T}_k, %\mathbf{R}_k, s_k) \nonumber \\
%                               &
%\xinke{what is morphological operation?} 
In order to remove noise points near the predict mask boundary of the object, we estimate the segmentation mask of the object $\boldsymbol{\Omega}_\mathrm{obj}$ by performing the morphological operations erosion on the predicted segmentation mask $\hat{\boldsymbol{\Omega}}$. Then, the point cloud of the object $\mathbf{P}_\mathrm{obj}$ can be computed by back-projecting the pixels in $\boldsymbol{\Omega}_\mathrm{obj}$:
\begin{equation}
    \mathbf{P}_\mathrm{obj} = \{\mathbf{D}(u, v)\mathbf{K}^{-1}(u, v, 1)^T, (u, v)\in \boldsymbol{\Omega}_\mathrm{obj}\},
\end{equation}
where $\mathbf{D}$ denotes the depth image, and $\mathbf{K}$ represents the intrinsic matrix of the camera.

Built upon PointNet++ \cite{qi2017pointnet++}, LatentNet takes the normalized point cloud as input and regresses a latent vector that encodes the 3D object shape:
\begin{equation}
    h_{\gamma}(\bar{\mathbf{P}}_{obj}) = \mathbf{z} \label{eq:latentnet},
\end{equation}
where $\gamma$ denotes the parameters of the LatentNet $h$.
This latent vector $\mathbf{z}$ can be decoded to a 3D shape represented by a continuous SDF as in \cite{park2019deepsdf}:
\begin{equation}  \label{eq:sdf}
    f_{\theta}(\mathbf{z}, \mathbf{x}) \approx SDF(\mathbf{x}),
\end{equation}
where the decoder $f_{\theta}$ with parameters $\theta$ takes in the latent code $\mathbf{z}$, a query 3D point $\mathbf{x}$ and outputs the SDF value of the shape at location $\mathbf{x}$. Given the decoding model $f_{\theta}$, the shape associated with latent vector $\mathbf{z}$ can be represented with the zero iso-surface of $f_{\theta}(\mathbf{z}, \mathbf{x})$. Thanks to the auto-decoder learning method, DeepSDF \cite{park2019deepsdf} maximizes the joint log posterior over all training shapes with respect to the individual shape code and the network parameters. Therefore, the DeepSDF network is trained, the optimized latent code can be used to approximate the ground truth for each object shape as well as used as the regression target for LatentNet. 

\subsection{Category-level Object Pose Refinement} \label{sec:refine}
Due to the limited number of particles in the particle filter for the time budget and the discretization of 3D rotation in each particle, we perform a continuous optimization using 3D points from the depth image $\mathbf{P}_\mathrm{obj}$ to refine the estimated pose. We optimize the pose by matching these points against the SDF estimated from Eq.~\eqref{eq:sdf} of the object. The optimization problem we solve is  

\begin{equation}
    (\hat{\mathbf{T}}, \hat{\mathbf{R}}) = \arg\min_{\mathbf{T},\mathbf{R}} \sum_{\mathbf{p}_i\in \mathbf{P}_\mathrm{obj}} |f_{\theta}(\mathbf{z}, g(\mathbf{p}_i, \mathbf{T}, \mathbf{R}, s))|, \label{eq:sdf_refine}
\end{equation}
where $\mathbf{p}_i$ is a 3D point in the point cloud $\mathbf{P}_\mathrm{obj}$ in the camera frame, the function $g$ is defined in Eq.~\eqref{eq:point_normalize} which transforms points in camera frame to object frame according to pose and size, and $f_{\theta}(\mathbf{z}, g(\mathbf{p}_i, \mathbf{T}, \mathbf{R}, s))$ represents the signed distance value estimated from Eq.~\eqref{eq:sdf} based on the shape latent $\mathbf{z}$.

% based on the current pose, size and shape latent code estimates. %\xinke{add more details of the auto-decoder network for computing the sdf values; cite deepsdf here.} 

Eq.~\eqref{eq:sdf_refine} indicates that improving the shape estimate $\mathbf{z}$ could improve SDF estimates and thus improving the refined poses. At the same time, a shape estimate relies directly on pose estimate, see Eq.~\eqref{eq:point_normalize}-\eqref{eq:latentnet}. Therefore, we perform shape estimation using Eq.~\eqref{eq:latentnet} given the estimated pose and perform pose refinement using Eq.~\eqref{eq:sdf_refine} given the estimated shape. For each image frame, we perform this “shape estimation” and “pose refinement” pair multiple times in an iterative way to further boost the estimation performance. At the same time, the shape estimation is also improved. %\xinke{replace good with improved? may need some rewording here.}  

%% file: 3_experiments.tex
\section{Experiments}
\subsection{Experimental Settings}
\textbf{Training Data Generation}.
The training data for both the auto-encoder and LatentNet is generated by rendering a ShapeNet object at random rotations. For the auto-encoder, the rendered images are superimposed at random crops of the MS-COCO dataset \cite{lin2014microsoft}. For LatentNet, the partial objects point clouds are obtained through back-projection from rendered depth images. We perturb the ground truth rotations by adding random Gaussian noises with $5\degree$ standard deviation. We then down-sample the point clouds to $N = 4873$ points using Furthest Point Sampling (FPS). 
The ground truth shape latent vector are generated by training DeepSDF~\cite{park2019deepsdf} using the default setting on ShapeNet objects. 
% at resolution $128\times128$. The background data for normalized depth is generated by averaging the RGB channels at random crops in MS-COCO dataset. 

%we need the paired data of normalized point clouds and their corresponding ground truth shape latent vector. 
%To train the LatentNet, we generate the synthetic data on-the-fly by rendering a ShapeNet object at random rotations. Then the partial objects point clouds are obtained through back-projection. We perturb the ground truth rotation by adding random Gaussian noise with $5\degree$ standard deviation. We then downsample the point clouds to $N = 4873$ points using Furthest Point Sampling (FPS). 

%Then, we generate the normalized point clouds. 
%The rendered depth images are distorted with multiplicative gamma noise and a few randomly dropped ellipses for network robustness.

% \textbf{Network Structure}.The auto-encoder takes in normalized depth map of size $128\times128$, and consists of four $5 \times5$ convolutional layers and four $5 \times 5$ deconvolutional layers with stride 2 for the encoder and the decoder, respectively. After the convolutional layers, a fully connected layer is used to produce 128 dimensional embeddings. 

% The LatentNet uses PointNet++ \cite{qi2017pointnet++} MSG classification network with two fully connected layer in the end. The dimension of the output predicted latent vector is 256.

\textbf{Training Details}.
Both the auto-encoders and the LatentNets are trained for each object category separately for $225,000$ iterations with batch size of $64$ using the Adam optimizer with a learning rate $0.0002$. They are optimized with the L2 loss with largest reconstruction errors. %\xinke{is the latentnet trained using the same loss?}

\textbf{Evaluation Datasets}. We evaluate our system on the NOCS-REAL275 dataset \cite{wang2019normalized}. The object categories in the dataset are: bottle, bowl, camera, can, laptop and mug. Three of them are categories with axes symmetry. There are in total six video sequences with 3,200 frames in the testing set, and for each category there are three unseen object instances. 

\begin{table*}
    \centering
    \caption{Category-level 6D object pose estimation results on NOCS-REAL275 dataset.}
    \label{tab:nocs tracking} 
    \begin{adjustbox}{width=0.9\textwidth}
    \begin{threeparttable}
    \begin{tabular}{c|c|c|c|c|c||c|c|c|c|c}
    \hline\hline
    \multicolumn{2}{c|}{Method}                    & NOCS~\cite{wang2019normalized}           & CASS~\cite{chen2020learning}         & FS-Net~\cite{chen2021fs}        & \textbf{Ours}           & ICP~\cite{Zhou2018}  & 6Pack~\cite{wang20196}        & CAPTRA~\cite{weng2021captra}         & BundleTrack~\cite{wen2021bundletrack}   & \textbf{Ours}         \\\hline
    \multicolumn{2}{c|}{Input}                     & RGBD          & RGBD           & RGBD           & RGBD           & Depth     & RGBD         & RGBD           & RGBD          & RGBD         \\\hline
    \multicolumn{2}{c|}{Setting}                   & \multicolumn{4}{c||}{Single frame}                                 & \multicolumn{5}{c}{Tracking}              \\\hline
    \multicolumn{2}{c|}{Initialization}            & -             & -              & -              & 2D det. \& seg.        & 6D GT   & Pert 6D GT           & Pert 6D GT           & Pert 6D GT            & \textbf{2D det. \& seg.}       \\\hline\hline
    \multirow{4}{*}{bottle}  & $5^\circ5\text{cm}$ & 5.5           & 18.56          & \textbf{42.19} & 16.05          & 10.1 & 24.5         & \textbf{79.46} & 86.5          & 16.95        \\
                             & IoU25              & 48.7          & 84.94          & -              & \textbf{99.05} & 29.9 & 91.1         & -              & \textbf{100}  & \textbf{100} \\
                             & Rot. err. (deg)    & 25.6          & \textbf{14.6}  & -              & 51.41          & 48   & 15.6         & 3.29           & 1.6           & 9.72         \\
                             & Trans. err. (cm)   & 14.4          & 18.39          & -              & \textbf{2.25}  & 15.7 & 4            & 2.6            & 2.3           & 2.2          \\\hline
    \multirow{4}{*}{bowl}    & $5^\circ5\text{cm}$ & \textbf{62.2} & 44.46          & 59.16          & 56.56          & 40.3 & 55           & 79.2           & \textbf{99.6}          & 70.37        \\
                             & IoU25              & 99.6          & 95.58          & -              & \textbf{99.86} & 79.7 & \textbf{100} & -              & 99.9          & 99.94        \\
                             & Rot. err. (deg)    & \textbf{4.7}  & 5.05           & -              & 13.89          & 19   & 5.2          & 3.5            & \textbf{1.7}           & 5.49         \\
                             & Trans. err. (cm)   & \textbf{1.2}  & 3.46           & -              & 1.7            & 4.7  & 1.7          & \textbf{1.43}           & 2.1           & 1.49         \\\hline
    \multirow{4}{*}{camera}  & $5^\circ5\text{cm}$ & 0.6           & 0.44           & 1.76           & \textbf{9.2}   & 12.6 & 10.1         & 0.41           & \textbf{85.8} & 9.32         \\
                             & IoU25              & 90.6          & 88.93          & -              & \textbf{98.9}  & 53.1 & 87.6         & -              & 99.9          & \textbf{100} \\
                             & Rot. err. (deg)    & 33.8          & \textbf{27.61} & -              & 48.71          & 80.5 & 35.7         & 17.82          & \textbf{3}    & 13.69        \\
                             & Trans. err. (cm)   & \textbf{3.1}  & 6.09           & -              & 3.12           & 12.2 & 5.6          & 35.53          & \textbf{2.1}  & 2.72         \\\hline
    \multirow{4}{*}{can}     & $5^\circ5\text{cm}$ & 7.1           & 32.98          & \textbf{40.55} & 19.87          & 17.2 & 22.6         & 64.7           & \textbf{99.2} & 45           \\
                             & IoU25              & 77            & 89.94          & -              & \textbf{95.1}  & 40.5 & 92.6         & -              & \textbf{100}  & 98.32        \\
                             & Rot. err. (deg)    & 16.9          & \textbf{8.85}  & -              & 53.23          & 47.1 & 13.9         & 3.43           & \textbf{1.5}  & 8.52         \\
                             & Trans. err. (cm)   & 4             & 5.5            & -              & \textbf{1.89}  & 9.4  & 4.8          & 5.69           & \textbf{2.1}  & 2.64         \\\hline
    \multirow{4}{*}{laptop}  & $5^\circ5\text{cm}$ & 25.5          & \textbf{38.22} & 16.59          & 9.46           & 14.8 & 63.5         & 94.03          & \textbf{99.9} & 26.07        \\
                             & IoU25              & 94.7          & \textbf{99.93} & -              & 75.39          & 50.9 & 98.1         & -              & \textbf{99.9} & 99.5         \\
                             & Rot. err. (deg)    & 8.6           & \textbf{8.25}  & -              & 11.18          & 37.7 & 4.7          & 2.24           & 1.5           & 8.76         \\
                             & Trans. err. (cm)   & \textbf{2.4}  & 3.5            & -              & 3.18           & 9.2  & 2.5          & \textbf{1.48}  & 2.2           & 3.19         \\\hline
    \multirow{4}{*}{mug}     & $5^\circ5\text{cm}$ & 0.9           & 1.45           & 8.74           & \textbf{22.56} & 6.2  & 24.1         & \textbf{55.17} & 53.6          & 21.82        \\
                             & IoU25              & 82.8          & 82.19          & -              & \textbf{99.96} & 27.7 & 95.2         & -              & 99.9          & \textbf{100} \\
                             & Rot. err. (deg)    & 31.5          & \textbf{29.49} & -              & 38.24          & 56.3 & 21.3         & 5.36           & \textbf{5.2}  & 10.69        \\
                             & Trans. err. (cm)   & 4             & 15.04          & -              & \textbf{1.65}  & 9.2  & 2.3          & \textbf{0.79}  & 2.2           & 1.31         \\\hline
    \multirow{4}{*}{overall} & $5^\circ5\text{cm}$ & 17            & 23.5           & \textbf{28.2}  & 22.28          & 16.9 & 33.3         & 62.16          & \textbf{87.4} & 31.59        \\
                             & IoU25              & 82.2          & 84.2           & \textbf{95.1}  & 94.63          & 47   & 94.2         & -              & \textbf{99.9} & 99.63        \\
                             & Rot. err. (deg)    & 20.2          & \textbf{15.64} & -              & 36.11          & 48.1 & 16           & 5.94           & \textbf{2.4}  & 9.48         \\
                             & Trans. err. (cm)   & 4.9           & 8.66           & -              & \textbf{2.3}   & 10.5 & 3.5          & 7.92           & \textbf{2.1}  & 2.26 \\       
    \hline\hline
    \end{tabular}
    \begin{tablenotes}[normal,flushleft]
      \item $^\dagger$For tracking, BundleTrack, 6-PACK, ICP require gound truth (GT) 6D pose for initialization. CAPTRA perturbed (Pert) the GT 6D pose for initialization. Our method only requires 2D detection and segmentation for particle sampling, indicating its effectiveness when the initial 6D pose is not available. 
      \item $^\dagger$The notation ``-‘’ means that the corresponding metric was not reported.
    \end{tablenotes}
    \end{threeparttable}
    \end{adjustbox}
    \vspace{-5mm}
\end{table*}

\textbf{Evaluation Metrics}. We follow the evaluation metrics in NOCS \cite{wang2019normalized} and report the following metrics: 1) $\mathbf{5^\circ5\textbf{cm}}$: the percentage of tracking results with the orientation error $< 5^\circ$ and translation error $< 5\text{cm}$; 2) \textbf{IoU25}: percentage of frames when the overlap between the ground truth 3D bounding box and the estimated 3D bounding box is greater than $25\%$; 3) $\mathbf{R}_\mathrm{err}$: mean rotation error in degrees, and 4) $\mathbf{T}_\mathrm{err}$: mean translation error in centimeters. 
Additionally, we employ the Chamfer Distance (CD) to evaluate shape reconstruction. 

\subsection{Results}

\begin{figure}
    \centering
    \includegraphics[width=0.45\textwidth]{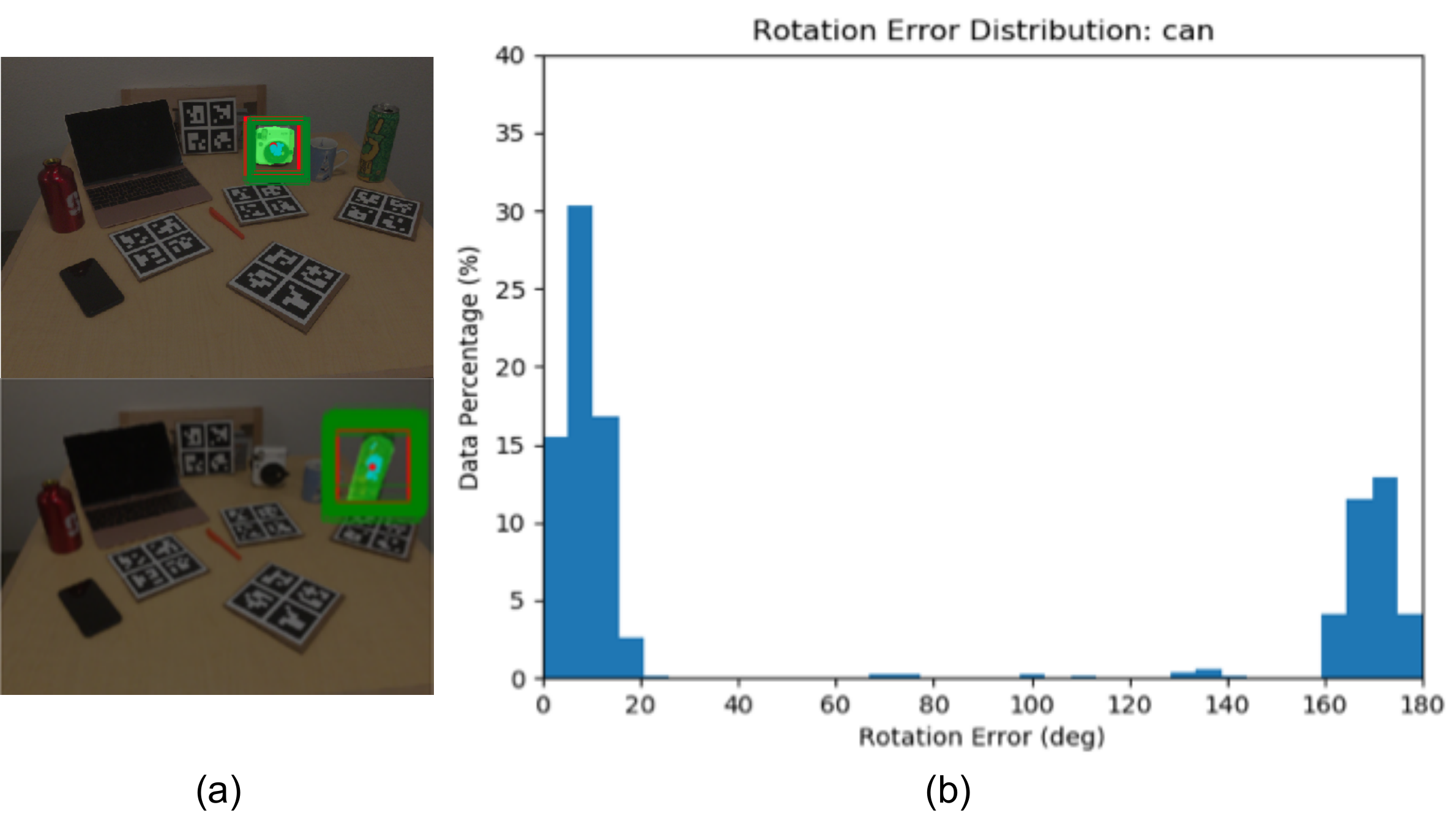}
    \caption{Rotation error analysis for single frame 6D pose estimation. (a) shows the superimposed RGB images with objects' model and the estimated 6D poses. Although the objects are well aligned, the rotation errors are $93^\circ$ and $179^\circ$ respectively due to the view symmetries from partial observations. We further visualize the rotation error distribution histogram for cans (b) to justify the large rotation errors are due to the view symmetry.}
    \label{fig:error_analysis}
    \vspace{-8mm}
\end{figure}

\textbf{Single Frame Category-level 6D Pose Estimation}. We first evaluate our system without exploiting temporal information. For every step, after initialization with 2D detection centers, we perform 10 filtering steps in the particle filtering. Then, shape estimation and pose refinement are performed. We compare our results with NOCS~\cite{wang2019normalized}, CASS~\cite{chen2020learning}, and FS-Net~\cite{chen2021fs} in Table~\ref{tab:nocs tracking}. For IoU25, our system is comparable to FS-Net and is significantly better than CASS and NOCS. The estimated translation is also significantly more accurate than CASS and NOCS. The rotation error is because for objects like \textit{bottle\_shengjun\_norm}, \textit{camera\_shengjun\_norm}, and the cans, the system fails to classify the objects' coordinate axes correctly from partial depth measurements, which results in around $90^\circ$ or $180^\circ$ rotation error due to the symmetric view. Some failure cases and a histogram of rotation errors are shown in Fig. \ref{fig:error_analysis}.

\begin{figure}
    \centering
         \includegraphics[width=0.48\textwidth]{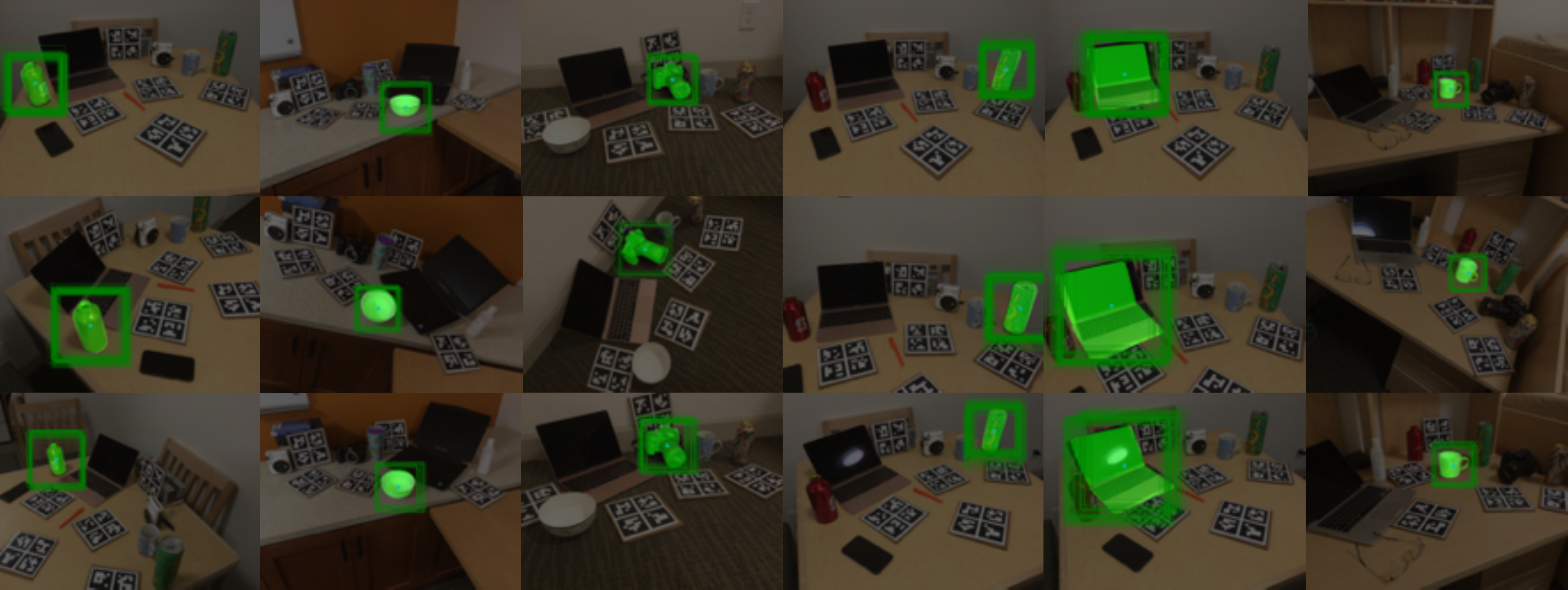}
    \caption{Visualization of estimated poses on the unseen objects on the NOCS-REAL275 dataset. The object models are superimposed on the images with the estimated poses. Green bounding boxes are the particle RoIs}
    \label{fig:pose visualization}
    \vspace{-4mm}
\end{figure}

\textbf{Category-level 6D Pose Tracking}. Our framework can also perform 6D pose tracking. The benefit of exploiting temporal consistency is obvious by comparing the single frame and tracking performance in Table~\ref{tab:nocs tracking}. We compare our category-level 6D object pose tracking system with ICP \cite{Zhou2018}, 6-PACK \cite{wang20196}, CAPTRA \cite{weng2021captra}, and BundleTrack \cite{wen2021bundletrack}. Overall, our system achieves better performance compared to 6-PACK and ICP in terms of overall IoU25, mean rotation error, and mean translation error. Comparing to CAPTRA which perturbs the ground truth 6D pose as initialization, we are competitive under translation error. In particular, we achieve much better performance on the challenging NOCS objects such as camera thanks to the SDF-based pose refinement. Although \cite{wen2021bundletrack} shows attractive performance at first glance, the requirement of using the ground truth 6D pose for initialization limits its application in the real world since it is not always easy to obtain the ground truth pose. Our system can be initialized with only 2D detection and segmentation, which makes it an useful alternative when the initial 6D pose is not available. 
The visualization of our pose estimation results is shown in Fig. \ref{fig:pose visualization}.

% Green bounding boxes are particle RoIs. The object models are superimposed on the input RGB images with estimated 6D poses.

%Although our tracking system is less accurate than \cite{weng2021captra} and \cite{wen2021bundletrack}, our system is able to be initialized with only 2D detection and segmentation in contrast to 6D pose in the other tracking systems, which makes our system an useful alternative when the initial 6D pose is not available. 

% \begin{figure*}
%     \centering
%     \includegraphics[width=0.65\textwidth]{figures/poses_visualization.pdf}
%     \caption{Visualization of estimated poses on the unseen objects on the NOCS-REAL275 dataset. The object models are superimposed on the images at the estimated poses.}
%     \label{fig:pose visualization}
%     \vspace{-4mm}
% \end{figure*}

\begin{table}
\centering
\caption{Evaluation of point cloud reconstruction accuracy with Chamfer Distance (CD) ($\times 10^{-3}$) metric.} \label{tab:shape}
\begin{adjustbox}{width=0.4\textwidth}
\begin{tabular}{c|c|c|c|c}
\hline \hline
  & CASS\cite{chen2020learning} & Shape-Prior \cite{tian2020shape} & FS-Net \cite{chen2021fs} & Ours\\
\hline \hline
bottle & 0.75 & 3.44  & 1.2  & \textbf{0.113}\\\hline
bowl & 0.38 & 1.21  & 0.39  & \textbf{0.043}\\\hline
camera & 0.77 & 8.89  & 0.44  & \textbf{0.272}\\\hline
can & 0.42 & 1.56  & 0.62  & \textbf{0.237}\\\hline
laptop & 3.73 & 2.91  & 2.23  & \textbf{0.882}\\\hline
mug & 0.32 & 1.02  & 0.29  & \textbf{0.131}\\\hline \hline
overall & 1.06 & 3.17  & 0.86  & \textbf{0.237}\\
\hline \hline
\end{tabular}
\end{adjustbox}
\vspace{-6mm}
\end{table}

\begin{figure}
    \centering
    \includegraphics[width=0.4\textwidth]{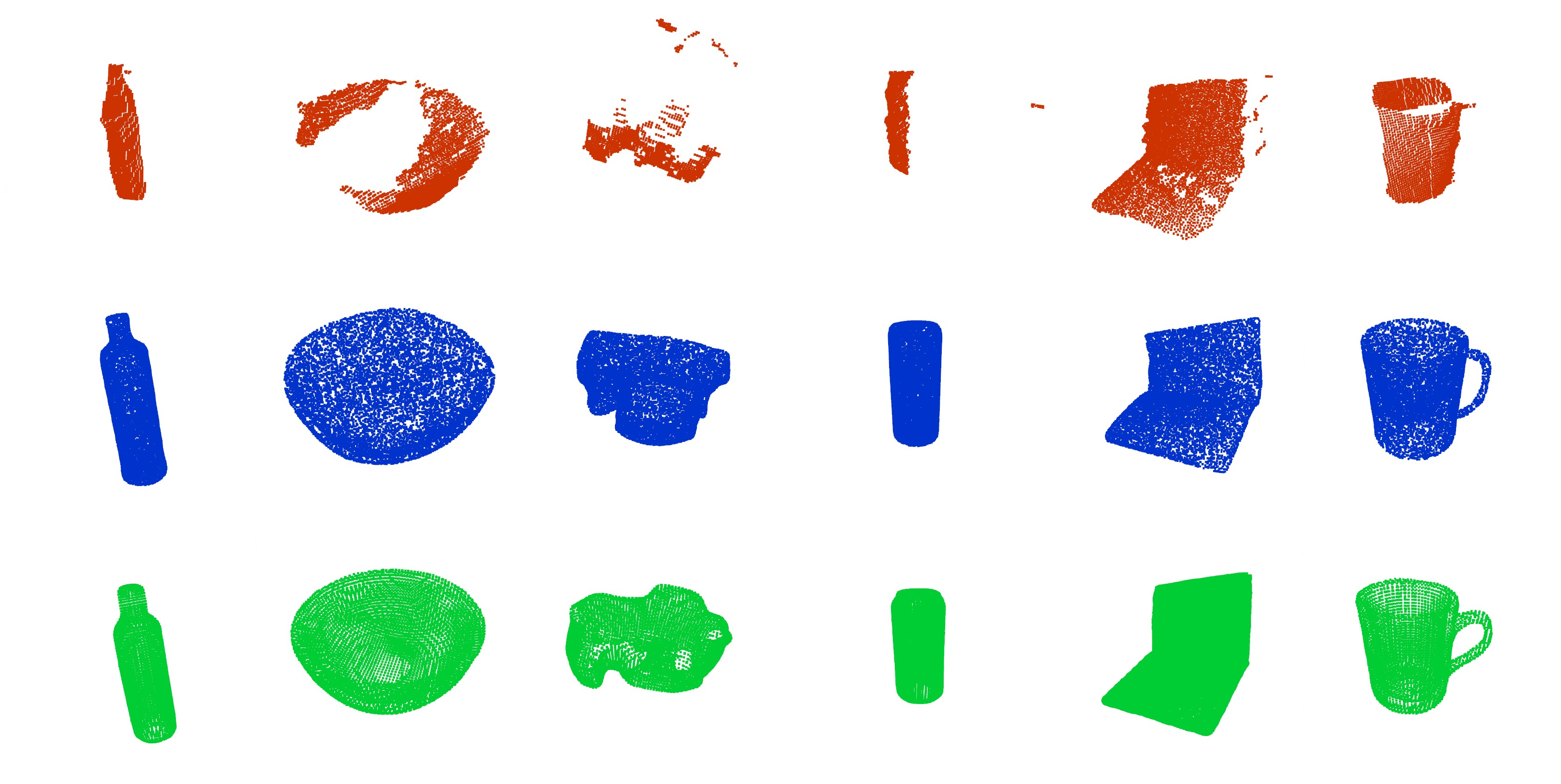}
    \caption{3D shape and size estimation visualization. The red, blue, and green point clouds represent the observations unprojected from depth maps, the predicted shapes, and the ground truth shapes respectively.}
    \label{fig:shape-estimation}
    \vspace{-4mm}
\end{figure}

\textbf{Shape and Size Estimation}. Table~\ref{tab:shape} reports a quantitative evaluation of 3D point cloud reconstruction from the estimated shape latent vector and the size. We compare our method with CASS \cite{chen2020learning}, Shape-Prior \cite{tian2020shape} and FS-Net \cite{chen2021fs}. Our method significantly outperforms those state-of-the-art methods thank to the accurate shape reconstruction with the implicit representation. We achieves an order of magnitude lower Chamfer distance comparing to CASS and Shape-Prior. Fig.~\ref{fig:shape-estimation} shows some qualitative visualization of the shape reconstruction. Our method is able to reconstruct full 3D shapes from the estimated shape latent vector, contrasting them with the point clouds back-projected from depth maps and the ground truth shape.

\subsection{Ablation Study}
An ablation study investigates the effectiveness of refinement module and the LatentNet is presented in Table~\ref{tab:architecture}.

We notice that without pose refinement, the estimation performance significantly deteriorates. However, the pipeline can achieve 12 fps running speed. Performing
refinement at some keyframe (every 10 frames) does slow down the pipeline but can boost the estimation performance. If performing refinement at every frame, the pipeline gets the best estimation performance with running speed is about 2 fps. Depending on the user’s application, there is a trade-off between the performance and the speed.

We also investigate the design choice for LatentNet by randomly initializing the shape latent for pose refinement. Without shape prediction, the estimation performance decreases significantly, indicating the effectiveness of LatentNet in assisting pose refinement. In addition, we compare the inference time of the LatentNet with the reconstruction time DeepSDF needs for shape estimation. Results show that LatentNet achieves higher efficiency: for example, in the bottle class, the inference time of LatentNet is 0.017s, while the DeepSDF requires 5.20s for shape reconstruction.

\begin{table}
\centering
\caption{Ablation Study on iCaps Architecture} \label{tab:architecture}
\begin{adjustbox}{width=0.45\textwidth}
\begin{tabular}{c|c|c|c|c|c}
\hline \hline
  & $5^\circ5\text{cm}$ & IoU25 & Rot. err. (deg) & Rot. err. (deg) & FPS\\
\hline \hline
Proposed & 31.59 & 99.63  & 9.48  & 2.26 & 1.84\\\hline
w/o refinement & 18.50 & 96.58  & 11.90  & 2.17 & 12.22\\\hline
w/o LatentNet & 17.95 & 98.65  & 13.15  & 2.91 & 1.93\\\hline
Refine/10 frames & 22.77 & 99.36  & 10.97  & 2.03 & 11.26\\
\hline \hline
\end{tabular}
\end{adjustbox}
\vspace{-6mm}
\end{table}

We also conduct the following ablation studies to our model to justify the various design choices for several key parameters of the developed method:
\begin{itemize}
    %\item \textbf{Particle sampling strategy}. We evaluate the system performance under different particle sampling strategy: one with sampling half of the particles according to the detection center is not used (w/o detection prior), the other one with do not exploit the temporal consistency but initialize the system in every frame according to the 2D detection (w/o temporal consistency).
    \item \textbf{Particle number}. We initialize the particles by uniformly sampling the size and sampling the translations according to the 2D detection and depth values. We use 300 particles for initialization and then use $P$ particles for pose tracking. We evaluate the system performance under different particle numbers $P$ in the particle filter.
    \item \textbf{Keyframe refinement}. We evaluate the system performance by performing the shape estimation and pose refinement every $K$ frames.
    \item \textbf{Iterative shape estimation and pose refinement}. We evaluate the system performance by performing the shape estimation and pose refinement in an iterative way $R$ times for one frame.
    \item \textbf{Pose refinement steps}. We evaluate the system performance by performing $F$ steps iterative optimization in one refinement.
\end{itemize}

\begin{figure}
    \centering
    \includegraphics[width=0.45\textwidth]{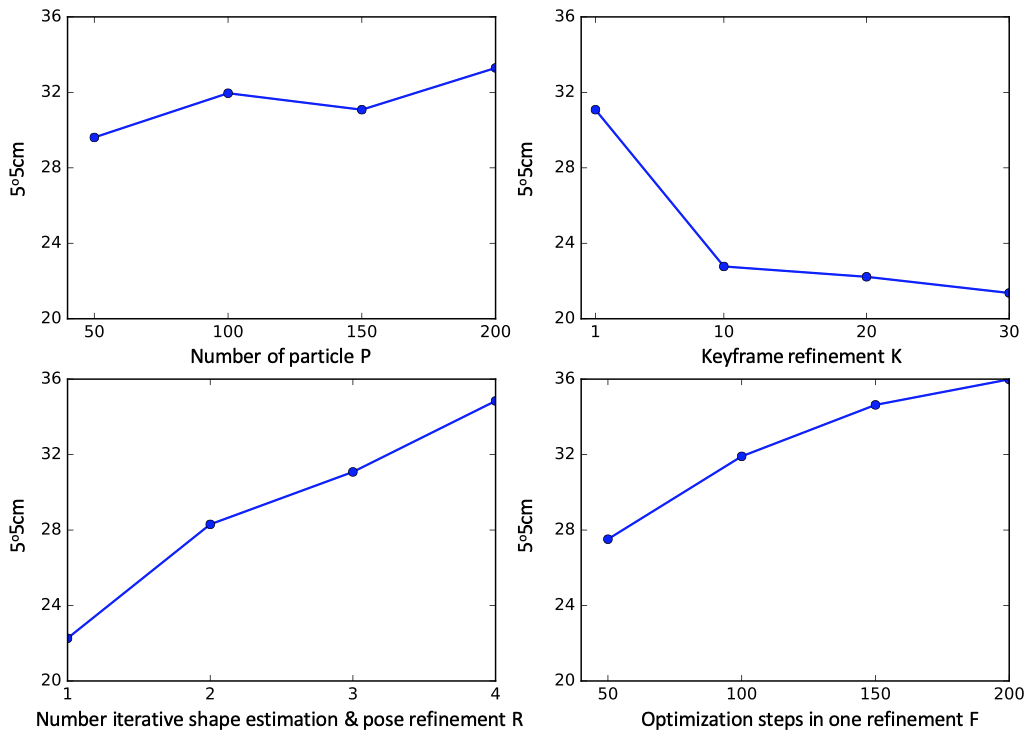}
    \caption{Ablation study of our approach.}
    \label{fig:ablation}
    \vspace{-7mm}
\end{figure}
%Table~\ref{tab:category-ablation} shows the ablation of particle sampling strategy. As can be seen in rows 1 and 3, sampling half of the particles according to the detection center results in significantly better detection performance. This is due to the abrupt motion and frames dropping in the NOCS-REAL275 dataset which can cause tracking failure if only the motion model is used. The detection center provides global information of the target object and significantly improves the robustness of category-level 6D object pose tracking. The results in rows 2 and 3 demonstrate the benefits of exploiting temporal consistency in tracking.

We reported the $5^\circ5\text{cm}$ performance of our approach under the rest of proposed ablations. From Fig.~\ref{fig:ablation}, we can see that as the particle number increases, the system performance improves. Pose refinement improves the estimation accuracy. Performing refinement more frequently contributes the high-precision estimation. Combining shape estimation and pose refinement in an iterative ways boosts the system performance since good shape leads to good pose estimation. At the same time, good shape estimates benefits from good pose estimation. In the pose refinement process, refinement step is a key parameter. However, too many particles or refinement steps requires heavy computation consumption and thus slow down the processing speed. Properly selecting all factors together contributes the high-precision pose estimation.

% \begin{table} \setlength{\tabcolsep}{4pt}
%     \centering
%     \caption{Ablation study on particle sampling strategy.}
%     \label{tab:category-ablation}
%     \begin{tabular}{c|c|c}
%         \hline\hline
%         Rows & Model                                                                      & IoU25 \\ \hline\hline
%         1    & \begin{tabular}[c]{@{}c@{}}Ours \\ (w/o detection prior)\end{tabular}      & 94.57 \\ \hline
%         2    & \begin{tabular}[c]{@{}c@{}}Ours \\ (w/o temporal consistency)\end{tabular} & 93.69 \\ \hline
%         3    & Ours                                                                       & 99.63 \\ \hline\hline
%         \end{tabular}
% \end{table}

% We also observe that the rotation error frequently appears between $5^\circ$ and $15^\circ$ by comparing the rotation error distributions between 6-PACK~\cite{wang20196} and our method in Figs. \ref{fig:rot error dist bottle} to \ref{fig:rot error dist mug}, which explains why the performance on $5^\circ5\text{cm}$ is not as good. 